\documentclass[final]{siamltex}
\usepackage{graphicx}
\usepackage{mathtools}
\usepackage{latexsym}
\usepackage[mathscr]{euscript}
\usepackage{xcolor}
\usepackage{algorithm,algorithmic}
\usepackage{fancyhdr}
\usepackage{multirow}
\usepackage{epsfig}
\usepackage{mathptmx}
\usepackage[most]{tcolorbox}
\usepackage{helvet}
\usepackage{courier}
\usepackage{type1cm}
\usepackage{imakeidx}
\usepackage[a4paper,textheight=23cm,textwidth=15cm]{geometry}
\usepackage{hyperref}
\usepackage{enumitem}
\usepackage{titlesec}
\usepackage{amsmath,amssymb,amsfonts}
\usepackage{tikz}
\usepackage{bm}
\usepackage{booktabs}
\usetikzlibrary{arrows.meta,shapes,positioning,calc,fit,backgrounds,decorations.pathreplacing}
\usepackage{pifont}

\newtheorem{remark}{Remark}

\newcommand{\nd}{\noindent}

\title{A Computationally Efficient Multidimensional Vision Transformer}
	\author{A. Elichi\thanks{Université du Littoral Cote d'Opale, LMPA, 50 rue F. Buisson, 62228 Calais-Cedex, France.}   \and K. Jbilou\footnotemark[1]  \and  F. Dufrenois\thanks{ LISIC, 50 rue F. Buisson, Universit\'e du Littoral Cote d'Opale, 62228 Calais-Cedex, France.}} 

\date{}

\begin{document}
	\maketitle
	
	\begin{abstract}
		Vision Transformers have achieved state-of-the-art performance in a wide range
		of computer vision tasks, but their practical deployment is limited by high
		computational and memory costs. In this paper, we introduce a novel tensor-based
		framework for Vision Transformers built upon the \emph{Tensor Cosine Product}
		(c-product). By exploiting multilinear structures inherent in image data and the
		orthogonality of cosine transforms, the proposed approach enables efficient
		attention mechanisms and structured feature representations. We develop the
		theoretical foundations of the tensor cosine product, analyze its algebraic
		properties, and integrate it into a new c-product-based Vision Transformer
		architecture (TCP-ViT). Numerical experiments on standard classification and
		segmentation benchmarks demonstrate that the proposed method achieves a uniform
		$1/C$ parameter reduction (where $C$ is the number of channels) while
		maintaining competitive accuracy.
		
	\end{abstract}

	\section{Introduction}
	\label{sec:introduction}
	
	Vision Transformers (ViTs)~\cite{dosovitskiy2021} have recently attracted
	considerable interest as a strong alternative to convolutional neural networks,
	owing to their ability to capture long-range dependencies via self-attention.
	Unlike convolutional architectures, which operate on local receptive fields with
	shared weights, ViTs facilitate global information exchange across the entire
	image from the earliest network layers. This global modeling capability has
	enabled ViTs to achieve state-of-the-art performance across a wide range of
	computer vision tasks, including image classification, object detection, and
	semantic segmentation.
	
	Despite these successes, conventional ViT models suffer from several intrinsic
	limitations. Images are first divided into patches that are subsequently
	flattened into vectors and processed as independent tokens. This vectorization
	step disrupts the inherent multidimensional structure of image data and neglects
	correlations across spatial and channel dimensions. In addition, the
	self-attention mechanism exhibits quadratic computational and memory complexity
	with respect to the number of tokens, rendering ViTs costly to train and deploy
	on high-resolution images. These factors significantly constrain their
	scalability and practicality in resource-limited environments.
	
	To alleviate these issues, extensive research has been devoted to the
	development of efficient Vision Transformer variants. Proposed approaches
	include sparse and axial attention mechanisms, low-rank and kernel-based
	attention formulations, as well as hybrid architectures that combine
	convolutional and transformer
	components~\cite{touvron2021training,wang2021axial,guo2021cmt}. Although
	these methods reduce computational overhead, they generally retain vectorized
	representations and do not explicitly exploit the multilinear structure of
	visual data. Consequently, important spatial and spectral relationships may
	remain underutilized.
	
	Recent advances in tensor numerical linear algebra provide an alternative and
	mathematically principled approach to addressing these challenges. Tensors offer
	natural representations for multidimensional data and support algorithms that
	preserve structural properties while reducing
	redundancy~\cite{Hitchcock1927,Kilmer2011,Kolda2009}. In particular, tensor
	products based on orthogonal transforms, such as Fourier and cosine transforms,
	have demonstrated strong effectiveness in large-scale inverse problems,
	regularization of ill-posed systems, and high-dimensional scientific
	computing~\cite{bai2020multilinear,ElGuideJbilouSadaka2021}. These techniques
	benefit from solid theoretical foundations and efficient implementations enabled
	by fast transform algorithms.
	
	Motivated by these developments, we propose an efficient Vision Transformer
	framework based on the \emph{tensor cosine product} (c-product). By operating
	directly on tensor-valued representations and employing cosine-transform-based
	tensor products, the proposed method preserves spatial structure, reduces
	computational complexity, and avoids the use of complex-valued arithmetic. The
	cosine transform is particularly well suited for vision applications due to its
	strong energy compaction properties and favorable boundary
	behavior~\cite{martin2018cosine}.
	
	The tensor cosine product enables a structured formulation of attention
	mechanisms in which interactions between image patches are computed within a
	multilinear transform domain. This formulation leads to more efficient attention
	computation, improved memory utilization, and a natural integration of numerical
	linear algebra techniques into modern deep learning
	architectures~\cite{falcao2022efficient}.
	
	The main contributions of this work are summarized as follows:
	\begin{itemize}[leftmargin=2em]
		\item We formally define the tensor cosine product and establish its
		fundamental algebraic and orthogonality-preserving properties.
		\item We introduce a c-product-based self-attention mechanism that
		preserves tensor structure while reducing computational and memory
		complexity.
		\item We design a cosine-product Vision Transformer (TCP-ViT) that operates
		directly on tensor-valued image patches, achieving a uniform $1/C$
		parameter reduction across all linear components.
		\item We provide theoretical complexity analysis and numerical experiments
		demonstrating the efficiency and competitiveness of the proposed approach.
	\end{itemize}
	
	The remainder of this paper is structured as follows.
	Section~\ref{sec:related} reviews related work on efficient Vision Transformers
	and tensor methods in deep learning.
	Section~\ref{sec:cosine_product} introduces the discrete cosine transform and
	the tensor cosine product with its algebraic properties.
	Section~\ref{sec:std_vit} recalls the standard Vision Transformer architecture
	and establishes the notation used throughout.
	Section~\ref{sec:cvit} presents the tensorized TCP-ViT architecture.
	Section~\ref{sec:analysis} provides the parameter efficiency and computational
	complexity analyses.
	Section~\ref{sec:experiments} reports numerical experiments.
	Section~\ref{sec:conclusion} concludes the paper.

	\section{Related Work}
	\label{sec:related}
	
	\subsection{Efficient Vision Transformers.}
	The original ViT~\cite{dosovitskiy2021,Khan2022} demonstrated that pure transformer
	architectures can match or exceed CNNs on image classification when trained on
	large-scale datasets. However, its quadratic attention cost motivated numerous
	efficient variants. DeiT~\cite{touvron2021training} introduced knowledge
	distillation and data-efficient training strategies.
	Swin Transformer~\cite{liu2021swin} employs shifted window attention to achieve
	linear complexity in the number of tokens.
	Axial attention~\cite{wang2021axial} factorizes the full attention along
	spatial axes. These approaches reduce computational cost but retain the
	vectorized token representation and do not explicitly leverage the multilinear
	structure of image data.
	
	\subsection{Tensor Methods in Deep Learning.}
	Tensor decompositions have been widely applied to compress neural networks.
	Tucker decomposition~\cite{Kim2016tucker} and tensor-train
	(TT)~\cite{novikov2015tensorizing} formats have been used to factorize weight
	matrices in fully connected and convolutional layers, achieving significant
	parameter reductions. LASER~\cite{sharma2024laser} showed that applying SVD to
	individual weight matrices and removing components corresponding to the
	smallest singular values can improve LLM reasoning, particularly when
	compressing the feed-forward network. More recently, the t-product
	framework~\cite{Kilmer2011,Kilmer2013} has enabled structured tensor operations
	based on the discrete Fourier transform (DFT), with applications to tensor
	regression, principal component analysis~\cite{Hached2021,Dufrenois2023}, and
	inverse problems~\cite{Elhachimi2024,ElGuideJbilouSadaka2021}. The present work
	adopts the cosine-transform variant of the t-product, which operates entirely in
	the real domain and avoids the complex arithmetic inherent to DFT-based methods.
	
	\subsection{Tensor-based Attention Compression.}
	The most closely related work is TensorLLM~\cite{gu2025tensorllm}, which
	tensorises multi-head attention weights by stacking the per-head query, key,
	value, and output matrices into a 4D tensor
	and applying Tucker decomposition with shared factor matrices across heads.
	While both our approach and TensorLLM exploit the multi-head structure of
	attention, the two methods differ fundamentally in three respects.
	First, TensorLLM is a \emph{post-hoc} compression technique: it decomposes
	pre-trained weights via an approximate low-rank factorization, which
	introduces approximation error controlled by the chosen multilinear ranks
	In contrast, our c-product formulation is a \emph{native
		algebraic reformulation}: the model is trained directly with the c-product
	, and the resulting factorization is
	\emph{exact}, with parameter reduction arising
	from the block-diagonal structure rather than from rank truncation.
	Second, TensorLLM requires hyperparameter selection for the multilinear ranks
	and produces variable compression ratios depending on the chosen ranks, whereas
	the c-product yields a \emph{uniform, architecture-independent}  
	parameter reduction that is determined solely by the number of channels.
	Third, the cross-channel coupling mechanism differs: TensorLLM learns shared
	factor matrices at additional computational cost, while the c-product
	couples channels \emph{implicitly} through the fixed orthogonal DCT matrix
	$\Phi_C$ at zero parametric cost.

    In a related but distinct direction, Su et al.~\cite{su2024dctvit} proposed
DctViT, a hybrid CNN--Transformer architecture that employs the DCT for
feature map compression within a dedicated convolutional block (DAD block).
Their approach applies the DCT as a signal processing tool to reduce spatial
redundancy in intermediate representations, while retaining a standard
Transformer encoder with conventional matrix multiplications. In contrast,
the present work uses the DCT as the algebraic foundation of a tensor
product ($\star_c$) that replaces \emph{all} linear projections in the
Transformer encoder, yielding a principled and uniform $1/C$ parameter
reduction across the entire architecture without any convolutional component.

	\section{The Cosine Product and Its Properties}
	\label{sec:cosine_product}
	
	We begin by introducing the notation and definitions that will be used
	throughout the paper.
	
	\subsection{Notation}
	\label{sec:notation}
	
	Scalars are denoted by lowercase letters ($a$), vectors by bold lowercase
	letters ($\mathbf{x}$), matrices by uppercase letters ($X$), and third-order
	tensors by calligraphic letters ($\mathcal{X}$). The notation and symbols used
	throughout this paper are summarized in Table~\ref{tab:notation}.
	
	\begin{table}[htbp]
		\centering
		\caption{Summary of notation used throughout the paper.}
		\label{tab:notation}
		\renewcommand{\arraystretch}{1.3}
		\begin{tabular}{cl}
			\toprule
			\textbf{Symbol} & \textbf{Description} \\
			\midrule
			$H_{\mathrm{img}}, W_{\mathrm{img}}$ & Image height and width \\
			$C$ & Number of image channels (tube dimension) \\
			$P$ & Patch side length \\
			$N = H_{\mathrm{img}} W_{\mathrm{img}} / P^2$ & Number of patches (tokens) \\
			$d = P^2$ & Spatial embedding dimension per frontal slice \\
			$d_{\mathrm{eff}} = d \cdot C$ & Effective (flattened) embedding dimension \\
			$H$ & Number of attention heads \\
			$d_h = d / H$ & Per-head dimension (TCP-ViT); $d_h^{\mathrm{std}} = d_{\mathrm{eff}} / H$ (Std-ViT) \\
            $r_{\mathrm{ff}}$ & FFN expansion ratio ($d_{\mathrm{ff}} = r_{\mathrm{ff}} \cdot d$) \\
			$L$ & Number of Transformer layers \\
			$\star_c$ & Tensor cosine product \\
			$\mathcal{X}^{(k)}$ & $k$-th frontal slice of $\mathcal{X}$ \\
			$\widehat{\mathcal{X}}^{(k)}$ & $k$-th frontal slice of $\mathrm{DCT}_3(\mathcal{X})$ \\
			$\mathcal{X}^{\top_c}$ & Tensor c-transpose (Definition~\ref{def:ctranspose}) \\
			\bottomrule
		\end{tabular}
	\end{table}
	
	A third-order tensor is denoted by $\mathcal{X} \in \mathbb{R}^{m \times n \times C}$, where $C$ denotes the size along the third mode (the tube dimension). The $k$-th frontal slice is written $\mathcal{X}^{(k)} \in \mathbb{R}^{m \times n}$ for $k = 1, \ldots, C$.

	\subsection{Discrete Cosine Transform}
	\label{sec:dct}
	
	The orthogonal DCT-II matrix $\Phi_C \in \mathbb{R}^{C \times C}$ has entries
	defined by~\cite{golubvanloan}
	\begin{equation}\label{eq:dct_matrix}
		(\Phi_C)_{j,k} =
		\begin{cases}
			\sqrt{\dfrac{1}{C}}, & j = 0, \\[3mm]
			\sqrt{\dfrac{2}{C}} \, \cos\!\left[ \dfrac{\pi (2k+1) j}{2 C} \right], & j = 1,2,\dots,C-1,
		\end{cases}
	\end{equation}
	for $j,k = 0,1,\dots,C-1$.
	It satisfies the orthogonality property
	\begin{equation}\label{eq:dct_orthogonal}
		\Phi_C^\top \Phi_C = I_C, \qquad \Phi_C \Phi_C^\top = I_C.
	\end{equation}
	
	For a third-order tensor $\mathcal{X} \in \mathbb{R}^{m \times n \times C}$,
	the DCT along the third mode is defined as
	\begin{equation}\label{eq:dct_tensor}
		\widehat{\mathcal{X}} = \mathrm{DCT}_3(\mathcal{X}) = \mathcal{X} \times_3 \Phi_C,
	\end{equation}
	where $\times_3$ denotes the mode-3 product. The $k$-th frontal slice of the
	transformed tensor is denoted $\widehat{\mathcal{X}}^{(k)}$ for
	$k = 1, \ldots, C$. The inverse transform is
	\begin{equation}\label{eq:idct_tensor}
		\mathcal{X} = \mathrm{IDCT}_3(\widehat{\mathcal{X}}) = \widehat{\mathcal{X}} \times_3 \Phi_C^\top.
	\end{equation}
	
	The DCT possesses several properties that make it particularly suitable for
	tensor-based vision transformers:
	\begin{itemize}[leftmargin=2em]
		\item \emph{Energy compaction:} Most of the signal energy is concentrated
		in the low-frequency coefficients, enabling effective low-rank
		approximation and compression.
		\item \emph{Real-valued orthogonality:} Unlike the Fourier transform, the
		DCT produces real-valued orthogonal coefficients, simplifying
		computations in neural networks.
		\item \emph{Fast computation:} Using the fast cosine transform, the DCT
		can be computed in $O(C \log C)$ operations per tube, and
		$O(m n C \log C)$ for a third-order tensor along the third mode.
		\item \emph{Invertibility:} The DCT is perfectly invertible, ensuring no
		information loss when transforming between the spatial and cosine
		domains.
	\end{itemize}

	\subsection{Tensor Cosine Product}
	\label{sec:cprod}
	
	\begin{definition}[Tensor Cosine Product]\label{def:cprod}
		Let $\mathcal{A} \in \mathbb{R}^{m \times n \times C}$ and
		$\mathcal{B} \in \mathbb{R}^{n \times \ell \times C}$. The
		\emph{tensor cosine product} (c-product) is defined as
		\begin{equation}\label{eq:cprod}
			\mathcal{C} = \mathcal{A} \star_c \mathcal{B}
			= \mathrm{IDCT}_3\!\Big(
			\mathrm{DCT}_3(\mathcal{A}) \cdot \mathrm{DCT}_3(\mathcal{B})
			\Big)
			\;\in\; \mathbb{R}^{m \times \ell \times C},
		\end{equation}
		where the matrix product ``$\,\cdot\,$'' is performed slice-wise in the
		transform domain:
		$\widehat{\mathcal{C}}^{(k)} = \widehat{\mathcal{A}}^{(k)} \, \widehat{\mathcal{B}}^{(k)}$
		for $k = 1, \ldots, C$.
	\end{definition}
	
	This product allows efficient computation via fast DCT algorithms,
	orthogonality-preserving operations for numerical stability, and structured
	low-rank approximations.

	\subsection{Basic Tensor Operations}
	\label{sec:tensor_ops_basic}
	
	We summarize the fundamental definitions of third-order tensor operations
	under the c-product framework, extending the t-product algebra introduced
	by Kilmer et al.~\cite{Kilmer2011}.
	
	\begin{definition}[Tensor c-transpose]\label{def:ctranspose}
		Let $\mathcal{A} \in \mathbb{R}^{m \times n \times C}$. The
		\emph{c-transpose} $\mathcal{A}^{\top_c} \in \mathbb{R}^{n \times m \times C}$
		is the tensor whose frontal slices in the DCT domain satisfy
		\begin{equation}\label{eq:ctranspose}
			\big(\widehat{\mathcal{A}^{\top_c}}\big)^{(k)}
			= \big(\widehat{\mathcal{A}}^{(k)}\big)^\top, \qquad k = 1, \ldots, C.
		\end{equation}
		Equivalently, $\mathcal{A}^{\top_c} = \mathrm{IDCT}_3\!\big(\mathrm{DCT}_3(\mathcal{A})^{\top_{\mathrm{slice}}}\big)$, where $\top_{\mathrm{slice}}$ transposes each frontal slice independently.
	\end{definition}
	
	\begin{definition}[Identity tensor, orthogonality, invertibility]\label{def:tensor_basic}
		Let $\mathcal{A} \in \mathbb{R}^{m \times m \times C}$. Then:
		\begin{enumerate}[label=(\roman*), leftmargin=2em]
			\item The \emph{identity tensor} $\mathcal{I}_{m,C} \in \mathbb{R}^{m \times m \times C}$ is defined by requiring that its first frontal slice equals the $m \times m$ identity matrix $I_m$ and all remaining frontal slices are zero.
			
			\emph{Property:} All frontal slices of $\widehat{\mathcal{I}}_{m,C} = \mathrm{DCT}_3(\mathcal{I}_{m,C})$ coincide with $I_m$.
			
			\item A tensor $\mathcal{Q} \in \mathbb{R}^{m \times m \times C}$ is \emph{c-orthogonal} if
			$\mathcal{Q}^{\top_c} \star_c \mathcal{Q} = \mathcal{Q} \star_c \mathcal{Q}^{\top_c} = \mathcal{I}_{m,C}$.
			
			\item A tensor $\mathcal{Q}$ is \emph{f-orthogonal} if each frontal slice $\widehat{\mathcal{Q}}^{(k)}$ of its DCT transform is an orthogonal matrix.
			
			\item A tensor $\mathcal{D} \in \mathbb{R}^{m \times m \times C}$ is \emph{f-diagonal} if all frontal slices of $\widehat{\mathcal{D}}$ are diagonal matrices.
			
			\item A tensor $\mathcal{A} \in \mathbb{R}^{m \times m \times C}$ is \emph{invertible} if there exists $\mathcal{A}^{-1} \in \mathbb{R}^{m \times m \times C}$ such that
			$\mathcal{A} \star_c \mathcal{A}^{-1} = \mathcal{I}_{m,C}$.
		\end{enumerate}
	\end{definition}
	
	\begin{definition}[f-symmetry and f-positive-definiteness]\label{def:fsym}
		A square tensor $\mathcal{A} \in \mathbb{R}^{m \times m \times C}$ is called
		\emph{f-symmetric} if all frontal slices
		$\widehat{\mathcal{A}}^{(k)}$ are symmetric matrices.
		It is \emph{f-positive-definite} (resp.\ f-positive-semidefinite) if each
		$\widehat{\mathcal{A}}^{(k)}$ is symmetric positive definite
		(resp.\ symmetric positive-semidefinite).
	\end{definition}
	
	These concepts enable a consistent generalization of classical matrix operations
	to third-order tensors while preserving properties such as orthogonality,
	symmetry, and positive definiteness in the DCT domain. They form the algebraic
	foundation for the TCP-ViT architecture developed in
	Section~\ref{sec:cvit}.

	\section{Standard Vision Transformer}
	\label{sec:std_vit}
	
	Before introducing the tensorized architecture, we recall the standard Vision
	Transformer~\cite{dosovitskiy2021} and establish the notation that will serve
	as the basis for the c-product generalization. All operations in this section
	are matrix-based and operate on vectorized patch representations.
	
	\subsection{Patch Embedding}
	\label{sec:std_patch}
	
	Let $\mathcal{I} \in \mathbb{R}^{H_{\mathrm{img}} \times W_{\mathrm{img}} \times C}$
	denote an input image with $C$ channels. The image is partitioned into
	$N = H_{\mathrm{img}} W_{\mathrm{img}} / P^2$ non-overlapping patches of size
	$P \times P$. Each patch is flattened into a vector of dimension
	$d_{\mathrm{eff}} = P^2 C$ and projected to an embedding space via a learnable
	matrix $W_E \in \mathbb{R}^{d_{\mathrm{eff}} \times d_{\mathrm{eff}}}$:
	\begin{equation}\label{eq:std_embed}
		X_0 = \bigl[\mathbf{x}_{\mathrm{cls}};\; \mathbf{x}_1;\; \ldots;\; \mathbf{x}_N\bigr] + E_{\mathrm{pos}}
		\;\in\; \mathbb{R}^{(N+1) \times d_{\mathrm{eff}}},
	\end{equation}
	where $\mathbf{x}_{\mathrm{cls}} \in \mathbb{R}^{d_{\mathrm{eff}}}$ is a
	learnable classification token and
	$E_{\mathrm{pos}} \in \mathbb{R}^{(N+1) \times d_{\mathrm{eff}}}$ is a
	learnable positional embedding.

	\subsection{Scaled Dot-Product Attention}
	\label{sec:std_attention}
	
	Given an input $X \in \mathbb{R}^{(N+1) \times d_{\mathrm{eff}}}$, the query,
	key, and value matrices are obtained via linear projections:
	\begin{equation}\label{eq:std_qkv}
		Q = X W_Q, \qquad K = X W_K, \qquad V = X W_V,
	\end{equation}
	where $W_Q, W_K, W_V \in \mathbb{R}^{d_{\mathrm{eff}} \times d_h}$ are
	learnable weight matrices and $d_h = d_{\mathrm{eff}} / H$ is the per-head
	dimension. The scaled dot-product attention is defined as
	\begin{equation}\label{eq:std_attn}
		\mathrm{Attention}(Q, K, V)
		= \mathrm{softmax}\!\left(\frac{Q K^\top}{\sqrt{d_h}}\right) V
		\;\in\; \mathbb{R}^{(N+1) \times d_h}.
	\end{equation}
	The softmax is applied row-wise, so that each row of the attention matrix
	sums to one.

	\subsection{Multi-Head Self-Attention}
	\label{sec:std_mhsa}
	
	With $H$ attention heads, the multi-head self-attention (MHSA) is
	\begin{equation}\label{eq:std_mhsa}
		\mathrm{MHSA}(X)
		= \bigl[O_1;\; \ldots;\; O_H\bigr] \, W_O
		\;\in\; \mathbb{R}^{(N+1) \times d_{\mathrm{eff}}},
	\end{equation}
	where each head output $O_h = \mathrm{Attention}(XW_{Q,h}, XW_{K,h}, XW_{V,h}) \in \mathbb{R}^{(N+1) \times d_h}$,
	the concatenation $[\cdot]$ is along the column dimension, and
	$W_O \in \mathbb{R}^{d_{\mathrm{eff}} \times d_{\mathrm{eff}}}$ is a learnable output
	projection.

	\subsection{Feed-Forward Network and Layer Normalization}
	\label{sec:std_ffn}
	
	Each Transformer layer applies a two-layer feed-forward network (FFN) with
	expansion ratio $r_{\mathrm{ff}}$ and GELU activation:
	\begin{equation}\label{eq:std_ffn}
		\mathrm{FFN}(X) = \phi(X W_1) \, W_2,
	\end{equation}
	where $W_1 \in \mathbb{R}^{d_{\mathrm{eff}} \times d_{\mathrm{ff}}}$,
	$W_2 \in \mathbb{R}^{d_{\mathrm{ff}} \times d_{\mathrm{eff}}}$,
	$d_{\mathrm{ff}} = r_{\mathrm{ff}} \cdot d_{\mathrm{eff}}$, and
	$\phi$ denotes the element-wise GELU activation.
	
	Layer normalization (LN) is applied with learnable scale and shift vectors
	$\gamma, \beta \in \mathbb{R}^{d_{\mathrm{eff}}}$.
	
	A full Transformer block maps $X_{\ell-1}$ to $X_\ell$ via:
	\begin{align}
		Y_\ell &= X_{\ell-1} + \mathrm{MHSA}\!\big(\mathrm{LN}(X_{\ell-1})\big),
		\label{eq:std_block_attn} \\
		X_\ell &= Y_\ell + \mathrm{FFN}\!\big(\mathrm{LN}(Y_\ell)\big).
		\label{eq:std_block_ffn}
	\end{align}

	\subsection{Parameter Count}
	\label{sec:std_params}
	
	For a single Transformer layer operating at dimension
	$\delta = d_{\mathrm{eff}}$, the parameter count (excluding biases) is:
	\begin{equation}\label{eq:std_param_count}
		\Theta_{\mathrm{layer}}^{\mathrm{Std}}(\delta)
		= \underbrace{4\,\delta^2}_{\text{MHSA: } W_Q, W_K, W_V, W_O}
		+ \underbrace{2\,r_{\mathrm{ff}}\,\delta^2}_{\text{FFN: } W_1, W_2}
		+ \underbrace{4\,\delta}_{\text{LN: } \gamma, \beta \times 2}
		= (4 + 2r_{\mathrm{ff}})\,\delta^2 + 4\,\delta.
	\end{equation}
	
	This $O(\delta^2)$ scaling is the fundamental cost that the c-product framework
	reduces by a factor of $C$, as shown in Section~\ref{sec:analysis}.

	\section{The TCP-ViT Architecture}
	\label{sec:cvit}
	
	We now lift each component of the standard ViT to the tensor setting by
	replacing matrix multiplication with the tensor cosine product $\star_c$. The
	key idea is to represent each patch as a third-order tensor
	$\mathcal{X}_i \in \mathbb{R}^{d \times C}$ (with $d = P^2$) rather than a
	flattened vector $\mathbf{x}_i \in \mathbb{R}^{d_{\mathrm{eff}}}$ (with
	$d_{\mathrm{eff}} = P^2 C$). All operations are then performed via the
	c-product, which couples the $C$ channels through the fixed orthogonal DCT at
	zero parametric cost.

	\subsection{Tensor Operations for Vision Transformers}
	\label{sec:tensor_ops}
	
	\begin{definition}[t-Linear]\label{def:tlinear}
		Given an input tensor $\mathcal{X} \in \mathbb{R}^{N \times d \times C}$ and a
		weight tensor $\mathcal{W} \in \mathbb{R}^{d \times d' \times C}$, the tensor
		linear projection is
		\begin{equation}\label{eq:tlinear}
			\mathrm{t\text{-}Linear}(\mathcal{X};\, \mathcal{W})
			= \mathcal{X} \star_c \mathcal{W}
			\;\in\; \mathbb{R}^{N \times d' \times C}.
		\end{equation}
	\end{definition}
	
	This operation generalizes the matrix multiplication $XW$ to the tensor
	setting. In the DCT domain, it decomposes into $C$ independent matrix
	multiplications:
	$\widehat{\mathcal{X}}^{(k)} \widehat{\mathcal{W}}^{(k)}$ for
	$k = 1, \ldots, C$. The cross-channel coupling is handled implicitly by the
	DCT/IDCT pair at zero parametric cost.
	
	\medskip
	
    \begin{definition}[t-Softmax]\label{def:tsoftmax}
    Given a tensor $\mathcal{S} \in \mathbb{R}^{N \times N \times C}$, the tensor
    softmax is defined slice-wise in the transform domain:
    \begin{equation}\label{eq:tsoftmax}
        \mathrm{t\text{-}Softmax}(\mathcal{S})_{ijk}
        = \frac{\exp(\widehat{\mathcal{S}}_{ijk})}
        {\displaystyle\sum_{\ell=1}^{N} \exp(\widehat{\mathcal{S}}_{i\ell k})},
        \qquad \forall\; i,k,
    \end{equation}
    where $\widehat{\mathcal{S}}^{(k)}$ denotes the $k$-th frontal slice in the
    DCT domain.
\end{definition}

The normalization is performed independently for each
token $i$ and each frontal slice~$k$ \emph{in the transform domain}. Since the
$C$ slices are decoupled in the DCT domain, the attention weights of each
frequency component are computed independently, yielding a block-diagonal
structure in the attention operator.  
	
	\medskip
	\begin{definition}[t-Attention]\label{def:tattn}
		Given query, key, and value tensors
		$\mathcal{Q}, \mathcal{K}, \mathcal{V} \in \mathbb{R}^{N \times d_h \times C}$,
		the tensor scaled attention is
		\begin{equation}\label{eq:tattn}
			\mathrm{t\text{-}Attention}(\mathcal{Q}, \mathcal{K}, \mathcal{V})
			= \mathrm{t\text{-}Softmax}\!\left(
			\frac{1}{\sqrt{d_h}}\;
			\mathcal{Q} \star_c \mathcal{K}^{\top_c}
			\right) \star_c \mathcal{V}
			\;\in\; \mathbb{R}^{N \times d_h \times C},
		\end{equation}
		where $\mathcal{K}^{\top_c} \in \mathbb{R}^{d_h \times N \times C}$ is the
		c-transpose (Definition~\ref{def:ctranspose}).
	\end{definition}
	
	
	\medskip
	\begin{definition}[t-MHSA]\label{def:tmhsa}
		Given an input $\mathcal{X} \in \mathbb{R}^{N \times d \times C}$, $H$
		attention heads with $d_h = d/H$, and learnable weight tensors
		\[
		\mathcal{W}_{Q,h},\; \mathcal{W}_{K,h},\; \mathcal{W}_{V,h}
		\in \mathbb{R}^{d \times d_h \times C},
		\qquad h = 1,\ldots,H,
		\]
		the tensor multi-head self-attention is
		\begin{equation}\label{eq:tmhsa}
			\mathrm{t\text{-}MHSA}(\mathcal{X})
			= \bigl[\mathcal{O}_1 ;\; \ldots ;\; \mathcal{O}_H\bigr]_{(2)}
			\star_c \mathcal{W}_O
			\;\in\; \mathbb{R}^{N \times d \times C},
		\end{equation}
		where $\mathcal{W}_O \in \mathbb{R}^{d \times d \times C}$,
		$[\cdot]_{(2)}$ denotes concatenation along mode~2, and each head output is
		\[
		\mathcal{O}_h = \mathrm{t\text{-}Attention}\!\big(
		\mathcal{X} \star_c \mathcal{W}_{Q,h},\;
		\mathcal{X} \star_c \mathcal{W}_{K,h},\;
		\mathcal{X} \star_c \mathcal{W}_{V,h}
		\big)
		\;\in\; \mathbb{R}^{N \times d_h \times C}.
		\]
	\end{definition}
	
	\noindent\textbf{Correspondence with the standard ViT.}
	In the standard ViT (Section~\ref{sec:std_mhsa}), each projection
	$W_{Q,h} \in \mathbb{R}^{d_{\mathrm{eff}} \times d_h}$ operates on the full
	flattened vector of dimension $d_{\mathrm{eff}} = dC$, requiring
	$d_{\mathrm{eff}} \cdot d_h = d C \cdot d_h$ parameters per head per
	projection. In the c-product formulation, the weight tensor
	$\mathcal{W}_{Q,h} \in \mathbb{R}^{d \times d_h \times C}$ contains only
	$d \cdot d_h \cdot C$ entries, but crucially these entries operate on the
	\emph{reduced} dimension $d$ per slice rather than the full dimension $dC$.
	This yields a $1/C$ parameter reduction per projection, as formalized in
	Section~\ref{sec:analysis}.
	
	\medskip
	
    \begin{definition}[t-LayerNorm]\label{def:tln}
    Given $\mathcal{X} \in \mathbb{R}^{N \times d \times C}$ and learnable tensors
    $\boldsymbol{\gamma}, \boldsymbol{\beta} \in \mathbb{R}^{1 \times d \times C}$,
    the tensor layer normalization is applied slice-wise in the transform domain:
    \begin{equation}\label{eq:tln}
        \mathrm{t\text{-}LN}(\widehat{\mathcal{X}})_{ijk}
        = \gamma_{1jk} \cdot
        \frac{\widehat{\mathcal{X}}_{ijk} - \widehat{\mu}_{ik}}
        {\widehat{\sigma}_{ik} + \varepsilon}
        + \beta_{1jk},
    \end{equation}
    where, for each token $i$ and frontal slice $k$ in the DCT domain:
    \[
    \widehat{\mu}_{ik} = \frac{1}{d}\sum_{j=1}^{d} \widehat{\mathcal{X}}_{ijk},
    \qquad
    \widehat{\sigma}_{ik} = \sqrt{\frac{1}{d}\sum_{j=1}^{d}
        (\widehat{\mathcal{X}}_{ijk} - \widehat{\mu}_{ik})^2}.
    \]
\end{definition}

\noindent Normalization is computed for each token--slice pair
$(i,k)$ independently in the transform domain, preserving the full tensor
structure. Since the DCT is an orthogonal transform,
$\|\Phi_C \mathbf{x}\|_2 = \|\mathbf{x}\|_2$, the norms of the activations
are preserved across domains, ensuring comparable conditioning in each slice.
	
	\medskip
	\begin{definition}[t-FFN]\label{def:tffn}
		Given $\mathcal{X} \in \mathbb{R}^{N \times d \times C}$ and weight tensors
		$\mathcal{W}_1 \in \mathbb{R}^{d \times d_{\mathrm{ff}} \times C}$,
		$\mathcal{W}_2 \in \mathbb{R}^{d_{\mathrm{ff}} \times d \times C}$ with
		$d_{\mathrm{ff}} = r_{\mathrm{ff}} \cdot d$, the tensor feed-forward network is
		\begin{equation}\label{eq:tffn}
			\mathrm{t\text{-}FFN}(\mathcal{X})
			= \phi\!\big(
			\mathcal{X} \star_c \mathcal{W}_1
			\big) \star_c \mathcal{W}_2
			\;\in\; \mathbb{R}^{N \times d \times C},
		\end{equation}
		where $\phi$ denotes the element-wise GELU activation.
	\end{definition}

	\subsection{Patch Tensorization}
	\label{sec:patch_tensor}
	
	Let $\mathcal{I} \in \mathbb{R}^{H_{\mathrm{img}} \times W_{\mathrm{img}} \times C}$
	be an input image. We partition it into $N = H_{\mathrm{img}} W_{\mathrm{img}} / P^2$
	non-overlapping patches and represent each patch as a matrix
	$X_i \in \mathbb{R}^{P^2 \times C}$, where $P^2 = d$ is the number of spatial
	positions within the patch and $C$ is the number of channels. The full set of
	patches is stacked as a third-order tensor:
	\begin{equation}\label{eq:patch}
		\mathcal{X} \in \mathbb{R}^{N \times d \times C},
		\qquad d = P^2.
	\end{equation}
	Mode~1 indexes the $N$ patches, mode~2 the $d$ spatial positions within each
	patch, and mode~3 the $C$ channels.
	
	\noindent\textbf{Comparison with the standard ViT.}
	The standard ViT flattens each patch to a vector
	$\mathbf{x}_i \in \mathbb{R}^{d_{\mathrm{eff}}}$ with
	$d_{\mathrm{eff}} = dC = P^2 C$, yielding a matrix
	$X \in \mathbb{R}^{N \times d_{\mathrm{eff}}}$. The TCP-ViT preserves the
	two-dimensional structure $d \times C$ within each patch, enabling the
	c-product to exploit cross-channel correlations algebraically.

	\subsection{Tensor Embedding}
	\label{sec:tensor_embed}
	
	A learnable classification tensor
	$\mathcal{T}_{\mathrm{cls}} \in \mathbb{R}^{1 \times d \times C}$ is prepended
	along mode~1, and a positional tensor
	$\mathcal{E} \in \mathbb{R}^{(N+1) \times d \times C}$ is added:
	\begin{equation}\label{eq:embed}
		\mathcal{X}_0
		= \bigl[\mathcal{T}_{\mathrm{cls}};\; \mathcal{X}\bigr]_{(1)} + \mathcal{E}
		\;\in\; \mathbb{R}^{(N+1) \times d \times C}.
	\end{equation}

	\subsection{TCP-ViT Transformer Block}
	\label{sec:cvit_block}
	
	Each of the $L$ TCP-ViT blocks maps
	$\mathcal{X}_{\ell-1} \in \mathbb{R}^{(N+1) \times d \times C}$ to
	$\mathcal{X}_\ell \in \mathbb{R}^{(N+1) \times d \times C}$ via:
	\begin{align}
		\mathcal{Y}_\ell
		&= \mathcal{X}_{\ell-1}
		+ \mathrm{t\text{-}MHSA}\!\big(
		\mathrm{t\text{-}LN}(\mathcal{X}_{\ell-1})\big),
		\label{eq:block_attn} \\[6pt]
		\mathcal{X}_\ell
		&= \mathcal{Y}_\ell
		+ \mathrm{t\text{-}FFN}\!\big(
		\mathrm{t\text{-}LN}(\mathcal{Y}_\ell)\big).
		\label{eq:block_ffn}
	\end{align}
	Every operation maps
	$\mathbb{R}^{(N+1) \times d \times C}$ to
	$\mathbb{R}^{(N+1) \times d \times C}$, ensuring that the tensor structure is
	preserved through all $L$ layers without any flattening or reshaping.

	\subsection{Algorithmic Description}
	\label{sec:algorithm}
	
	\begin{algorithm}[h]
		\caption{TCP-ViT Forward Pass}
		\label{alg:cvit}
		\begin{algorithmic}[1]
			\REQUIRE Image $\mathcal{I} \in \mathbb{R}^{H_{\mathrm{img}} \times W_{\mathrm{img}} \times C}$,
			$L$ blocks, weight tensors
			$\{\mathcal{W}_{Q,h}^\ell, \mathcal{W}_{K,h}^\ell,
			\mathcal{W}_{V,h}^\ell, \mathcal{W}_O^\ell,
			\mathcal{W}_1^\ell, \mathcal{W}_2^\ell\}_{\ell,h}$
			\STATE $\mathcal{X} \leftarrow \mathrm{Patchify}(\mathcal{I})
			\in \mathbb{R}^{N \times d \times C}$
			\hfill $\triangleright$ $d = P^2$, $N = H_{\mathrm{img}} W_{\mathrm{img}}/P^2$
			\STATE $\mathcal{X}_0 \leftarrow
			[\mathcal{T}_{\mathrm{cls}};\;\mathcal{X}]_{(1)} + \mathcal{E}$
			\hfill $\triangleright$ $\in \mathbb{R}^{(N{+}1) \times d \times C}$
			\FOR{$\ell = 1$ to $L$}
			\STATE $\bar{\mathcal{X}} \leftarrow \mathrm{t\text{-}LN}(\mathcal{X}_{\ell-1})$
			\hfill $\triangleright$ Normalize once, reuse for all heads
			\FOR{$h = 1$ to $H$}
			\STATE $\mathcal{Q}_h \leftarrow \bar{\mathcal{X}} \star_c \mathcal{W}_{Q,h}^\ell$
			\hfill $\triangleright$ $\in \mathbb{R}^{(N{+}1) \times d_h \times C}$
			\STATE $\mathcal{K}_h \leftarrow \bar{\mathcal{X}} \star_c \mathcal{W}_{K,h}^\ell$
			\STATE $\mathcal{V}_h \leftarrow \bar{\mathcal{X}} \star_c \mathcal{W}_{V,h}^\ell$
			\STATE $\mathcal{S}_h \leftarrow
			\frac{1}{\sqrt{d_h}}\;
			\mathcal{Q}_h \star_c \mathcal{K}_h^{\top_c}$
			\hfill $\triangleright$ $\in \mathbb{R}^{(N{+}1) \times (N{+}1) \times C}$
			\STATE $\mathcal{O}_h \leftarrow
			\mathrm{t\text{-}Softmax}(\mathcal{S}_h)
			\star_c \mathcal{V}_h$
			\hfill $\triangleright$ $\in \mathbb{R}^{(N{+}1) \times d_h \times C}$
			\ENDFOR
			\STATE $\mathcal{Y}_\ell \leftarrow
			\mathcal{X}_{\ell-1} +
			[\mathcal{O}_1;\ldots;\mathcal{O}_H]_{(2)}
			\star_c \mathcal{W}_O^\ell$
			\hfill $\triangleright$ Residual + output projection
			\STATE $\mathcal{H} \leftarrow
			\phi\!\big(
			\mathrm{t\text{-}LN}(\mathcal{Y}_\ell)
			\star_c \mathcal{W}_1^\ell\big)$
			\hfill $\triangleright$ $\in \mathbb{R}^{(N{+}1) \times d_{\mathrm{ff}} \times C}$
			\STATE $\mathcal{X}_\ell \leftarrow
			\mathcal{Y}_\ell +
			\mathcal{H} \star_c \mathcal{W}_2^\ell$
			\hfill $\triangleright$ $\in \mathbb{R}^{(N{+}1) \times d \times C}$
			\ENDFOR
			\RETURN $\mathcal{X}_\ell$
		\end{algorithmic}
	\end{algorithm}

	\begin{remark}[Lossless factorization]\label{rem:lossless}
		Unlike pruning, quantization, or low-rank approximation methods, the c-product
		factorization introduces no approximation error. Since
		$\Phi_C^\top \Phi_C = I_C$, the mapping
		$\mathcal{X} \mapsto \mathrm{IDCT}_3\!\big(\widehat{\mathcal{W}}^{(k)} \cdot
		\widehat{\mathcal{X}}^{(k)}\big)$ is exact for any choice of learned weights
		$\widehat{\mathcal{W}}^{(k)}$. The parameter reduction arises from structural
		constraints on the weight tensor (block-diagonal structure in the DCT domain),
		not from information discarding.
	\end{remark}

	\section{Theoretical Analysis}
	\label{sec:analysis}
	
	\subsection{Parameter Efficiency}
	\label{sec:param_efficiency}
	
	We compare the parameter counts of the TCP-ViT and the standard ViT. The TCP-ViT
	operates on tensors $\mathcal{X} \in \mathbb{R}^{N \times d \times C}$, while
	the standard ViT operates on the flattened representation
	$X \in \mathbb{R}^{N \times dC}$. Both architectures process the same amount
	of information per token.
	
	\subsubsection{Multi-Head Self-Attention.}
	With $H$ heads and per-head dimension $d_h = d/H$,
	Table~\ref{tab:mhsa_weights} compares the weight dimensions.
	
	\begin{table}[htbp]
		\centering
		\caption{MHSA weight dimensions. Each projection is listed per head;
			the total count sums over all $H$ heads.}
		\label{tab:mhsa_weights}
		\renewcommand{\arraystretch}{1.4}
		\begin{tabular}{lll}
			\toprule
			\textbf{Projection} & \textbf{TCP-ViT weight tensor} & \textbf{Std-ViT weight matrix} \\
			\midrule
			$\mathcal{W}_{Q,h}$ & $\mathbb{R}^{d \times d_h \times C}$ & $\mathbb{R}^{dC \times d_h C}$ \\
			$\mathcal{W}_{K,h}$ & $\mathbb{R}^{d \times d_h \times C}$ & $\mathbb{R}^{dC \times d_h C}$ \\
			$\mathcal{W}_{V,h}$ & $\mathbb{R}^{d \times d_h \times C}$ & $\mathbb{R}^{dC \times d_h C}$ \\
			$\mathcal{W}_O$      & $\mathbb{R}^{d \times d \times C}$ & $\mathbb{R}^{dC \times dC}$ \\
			\bottomrule
		\end{tabular}
	\end{table}
	
	\noindent\emph{TCP-ViT count:}
	The $H$ heads contribute $3H \cdot d \, d_h \, C = 3 d^2 C$ parameters
	(since $Hd_h = d$), plus $d^2 C$ for $\mathcal{W}_O$:
	\begin{equation}\label{eq:tcp_mhsa}
		\Theta_{\mathrm{MHSA}}^{\mathrm{c\text{-}ViT}} = 4\,d^2\,C.
	\end{equation}
	
	\noindent\emph{Standard count:}
	$3H \cdot (dC)(d_h C) = 3 d^2 C^2$, plus $(dC)^2 = d^2 C^2$ for $W_O$:
	\begin{equation}\label{eq:std_mhsa_count}
		\Theta_{\mathrm{MHSA}}^{\mathrm{Std}} = 4\,d^2\,C^2.
	\end{equation}
	
	\noindent\emph{Ratio:} $\Theta_{\mathrm{MHSA}}^{\mathrm{c\text{-}ViT}} \,/\,
	\Theta_{\mathrm{MHSA}}^{\mathrm{Std}} = 1/C$.
	
	\subsubsection{Feed-Forward Network.}
	With expansion ratio $r_{\mathrm{ff}}$ and hidden dimension
	$d_{\mathrm{ff}} = r_{\mathrm{ff}} \cdot d$,
	Table~\ref{tab:ffn_weights} compares the weight dimensions.
	
	\begin{table}[htbp]
		\centering
		\caption{FFN weight dimensions for TCP-ViT and standard ViT.}
		\label{tab:ffn_weights}
		\renewcommand{\arraystretch}{1.4}
		\begin{tabular}{lll}
			\toprule
			\textbf{Projection} & \textbf{TCP-ViT weight tensor} & \textbf{Std-ViT weight matrix} \\
			\midrule
			$\mathcal{W}_1$ & $\mathbb{R}^{d \times d_{\mathrm{ff}} \times C}$ & $\mathbb{R}^{dC \times d_{\mathrm{ff}} C}$ \\
			$\mathcal{W}_2$ & $\mathbb{R}^{d_{\mathrm{ff}} \times d \times C}$ & $\mathbb{R}^{d_{\mathrm{ff}} C \times dC}$ \\
			\bottomrule
		\end{tabular}
	\end{table}
	
	\noindent\emph{TCP-ViT count:}
	\begin{equation}\label{eq:tcp_ffn}
		\Theta_{\mathrm{FFN}}^{\mathrm{c\text{-}ViT}}
		= 2\,r_{\mathrm{ff}}\,d^2\,C.
	\end{equation}
	\noindent\emph{Standard count:}
	\begin{equation}\label{eq:std_ffn_count}
		\Theta_{\mathrm{FFN}}^{\mathrm{Std}}
		= 2\,r_{\mathrm{ff}}\,d^2\,C^2.
	\end{equation}
	\noindent\emph{Ratio:} $\Theta_{\mathrm{FFN}}^{\mathrm{c\text{-}ViT}} \,/\,
	\Theta_{\mathrm{FFN}}^{\mathrm{Std}} = 1/C$.
	
	\subsubsection{Layer Normalization.}
	Each t-LayerNorm has learnable tensors
	$\boldsymbol{\gamma}, \boldsymbol{\beta} \in \mathbb{R}^{1 \times d \times C}$.
	A TCP-ViT block uses two normalizations:
	\begin{equation}\label{eq:tcp_ln}
		\Theta_{\mathrm{LN}}^{\mathrm{c\text{-}ViT}} = 2 \times 2 \cdot dC = 4\,dC.
	\end{equation}
	The standard ViT has $\gamma, \beta \in \mathbb{R}^{dC}$ per normalization:
	\begin{equation}\label{eq:std_ln_count}
		\Theta_{\mathrm{LN}}^{\mathrm{Std}} = 2 \times 2 \cdot dC = 4\,dC.
	\end{equation}
	\noindent\emph{Ratio:} $1$ (normalization parameters scale linearly, so no
	quadratic gain).
	
	{Total per layer.}
	\begin{align}
		\Theta_{\mathrm{layer}}^{\mathrm{c\text{-}ViT}}
		&= (4 + 2r_{\mathrm{ff}})\,d^2\,C + 4\,dC,
		\label{eq:tcp_total} \\[4pt]
		\Theta_{\mathrm{layer}}^{\mathrm{Std}}
		&= (4 + 2r_{\mathrm{ff}})\,d^2\,C^2 + 4\,dC.
		\label{eq:std_total}
	\end{align}
	Over $L$ layers, the dominant quadratic terms yield:
	\begin{equation}\label{eq:param_ratio}
		\frac{\Theta^{\mathrm{c\text{-}ViT}}}{\Theta^{\mathrm{Std}}}
		\;\xrightarrow{d \gg 1}\;
		\frac{1}{C}.
	\end{equation}
	This result is independent of the depth $L$, the number of heads $H$, and the
	FFN expansion ratio $r_{\mathrm{ff}}$: the c-product framework yields a uniform
	$1/C$ parameter reduction across all linear components.
	
	\begin{table}[h!]
		\centering
		\caption{Per-layer parameter comparison between TCP-ViT and standard ViT.
			Input tensor $\mathcal{X} \in \mathbb{R}^{N \times d \times C}$, standard
			input $X \in \mathbb{R}^{N \times dC}$. Expansion ratio $r_{\mathrm{ff}}$,
			$H$ heads, $d_h = d/H$, $d_{\mathrm{ff}} = r_{\mathrm{ff}} d$.}
		\label{tab:param_general}
		\renewcommand{\arraystretch}{1.4}
		\begin{tabular}{lcccc}
			\toprule
			\textbf{Component}
			& \textbf{TCP-ViT tensor size}
			& \textbf{TCP-ViT params}
			& \textbf{Std params}
			& \textbf{Ratio} \\
			\midrule
			$\mathcal{W}_{Q,h}$ ($\times H$)
			& $\mathbb{R}^{d \times d_h \times C}$
			& $d^2 C$
			& $d^2 C^2$
			& $1/C$ \\
			$\mathcal{W}_{K,h}$ ($\times H$)
			& $\mathbb{R}^{d \times d_h \times C}$
			& $d^2 C$
			& $d^2 C^2$
			& $1/C$ \\
			$\mathcal{W}_{V,h}$ ($\times H$)
			& $\mathbb{R}^{d \times d_h \times C}$
			& $d^2 C$
			& $d^2 C^2$
			& $1/C$ \\
			$\mathcal{W}_O$
			& $\mathbb{R}^{d \times d \times C}$
			& $d^2 C$
			& $d^2 C^2$
			& $1/C$ \\
			\midrule
			\textbf{MHSA total}
			& ---
			& $4\,d^2\,C$
			& $4\,d^2\,C^2$
			& $\mathbf{1/C}$ \\
			\midrule
			$\mathcal{W}_1$
			& $\mathbb{R}^{d \times d_{\mathrm{ff}} \times C}$
			& $r_{\mathrm{ff}}\,d^2\,C$
			& $r_{\mathrm{ff}}\,d^2\,C^2$
			& $1/C$ \\
			$\mathcal{W}_2$
			& $\mathbb{R}^{d_{\mathrm{ff}} \times d \times C}$
			& $r_{\mathrm{ff}}\,d^2\,C$
			& $r_{\mathrm{ff}}\,d^2\,C^2$
			& $1/C$ \\
			\midrule
			\textbf{FFN total}
			& ---
			& $2r_{\mathrm{ff}}\,d^2\,C$
			& $2r_{\mathrm{ff}}\,d^2\,C^2$
			& $\mathbf{1/C}$ \\
			\midrule
			$\boldsymbol{\gamma}, \boldsymbol{\beta}$ ($\times 2$)
			& $\mathbb{R}^{1 \times d \times C}$ each
			& $4\,dC$
			& $4\,dC$
			& $1$ \\
			\midrule
			\textbf{Total per layer}
			& ---
			& $(4{+}2r_{\mathrm{ff}})\,d^2\,C$
			& $(4{+}2r_{\mathrm{ff}})\,d^2\,C^2$
			& $\boxed{1/C}$ \\
			\bottomrule
		\end{tabular}
	\end{table}
	
	The $1/C$ reduction originates from a fundamental algebraic property of the
	c-product. In the standard ViT, every learnable linear map operates on the
	full flattened vector of dimension $dC$, so that each weight matrix has
	$O(d^2 C^2)$ entries. The c-product framework replaces each such matrix by a
	third-order weight tensor $\mathcal{W} \in \mathbb{R}^{d \times d \times C}$
	with $O(d^2 C)$ entries. The cross-channel coupling that, in the standard
	setting, requires $O(C)$ additional parameters per connection is handled
	\emph{implicitly} by the fixed orthogonal DCT matrix $\Phi_C$ and its inverse
	$\Phi_C^\top$, which introduce zero learnable parameters. Since
	$\Phi_C^\top \Phi_C = I_C$, this factorization is exact: no information is
	lost, and no approximation is introduced.

	\subsection{Computational Complexity}
	\label{sec:flops}
	
	The FLOPs for a single TCP-ViT layer consist of $C$ independent matrix
	multiplications at dimension $d$, plus the DCT/IDCT overhead:
	\begin{equation}\label{eq:tcp_flops}
		F_{\mathrm{layer}}^{\mathrm{TCP\text{-}ViT}}
		= C \cdot F_{\mathrm{layer}}(N, d)
		+ O(N \, d \, C \log C),
	\end{equation}
	where $F_{\mathrm{layer}}(N, d) = (8d^2 + 2r_{\mathrm{ff}} d^2) N + 4N^2 d$
	is the FLOPs of a standard layer at dimension $d$. The standard layer operates
	at dimension $dC$:
	\begin{equation}\label{eq:std_flops}
		F_{\mathrm{layer}}^{\mathrm{Std}} = F_{\mathrm{layer}}(N, dC).
	\end{equation}
	

    Let $\alpha = (8 + 2r_{\mathrm{ff}})\,d$ denote the per-token projection
cost coefficient. The exact FLOPs ratio, neglecting the $O(C\log C)$ transform
overhead, is:
\begin{equation}\label{eq:flops_ratio}
    \frac{F^{\mathrm{c\text{-}ViT}}}{F^{\mathrm{Std}}}
    = \frac{\alpha + 4N}{\alpha\,C + 4N}.
\end{equation}
This ratio approaches $1/C$ in the projection-dominated regime
($\alpha \gg N$, i.e., large embedding dimension~$d$) and approaches~$1$ in
the attention-dominated regime ($N \gg \alpha$, i.e., long sequences).
	
	\begin{remark}[Attention complexity]\label{rem:attn_complexity}
		The attention computation within each frontal slice remains $O(N^2 d_h)$, i.e.,
		quadratic in the number of tokens $N$. The c-product framework reduces
		complexity with respect to the \emph{embedding dimension} (from $dC$ to $d$
		per slice), not the sequence length. This distinction is important: the TCP-ViT
		does not achieve sub-quadratic attention in $N$.
	\end{remark}

	\subsection{Training Considerations}
	\label{sec:training}
	
	

\nd In TCP-ViT, each learnable weight tensor
$\mathcal{W} \in \mathbb{R}^{d \times d' \times C}$ is updated through
the c-product structure. Consider a single t-Linear layer
(Definition~\ref{def:tlinear}):
$\mathcal{Y} = \mathcal{X} \star_c \mathcal{W}$, where
$\mathcal{X} \in \mathbb{R}^{N \times d \times C}$ and
$\mathcal{Y} \in \mathbb{R}^{N \times d' \times C}$.
By the chain rule applied to the c-product
(Definition~\ref{def:cprod}), the gradients of the loss~$\mathcal{L}$
with respect to the weight and input tensors are
\begin{align}
    \frac{\partial \mathcal{L}}{\partial \mathcal{W}}
    &= \mathcal{X}^{\top_c} \star_c
       \frac{\partial \mathcal{L}}{\partial \mathcal{Y}}
    \;\in\; \mathbb{R}^{d \times d' \times C},
    \label{eq:grad_w} \\[4pt]
    \frac{\partial \mathcal{L}}{\partial \mathcal{X}}
    &= \frac{\partial \mathcal{L}}{\partial \mathcal{Y}}
       \star_c \mathcal{W}^{\top_c}
    \;\in\; \mathbb{R}^{N \times d \times C},
    \label{eq:grad_x}
\end{align}
where $\mathcal{X}^{\top_c}$ and $\mathcal{W}^{\top_c}$ denote the
c-transposes (Definition~\ref{def:ctranspose}).
Since the c-product is built upon the orthogonal DCT matrix~$\Phi_C$.
Consequently, the Frobenius norms of the gradient tensors
\eqref{eq:grad_w}--\eqref{eq:grad_x} are identical in the spatial and
transform domains. The conditioning of the optimization landscape is
therefore preserved across all frontal slices, ensuring that no slice
suffers from vanishing or exploding gradients relative to the others.

\nd By the slice-wise property of the c-product
(Definition~\ref{def:cprod}), the $C$ frontal slices of the weight
tensor are updated independently in the transform domain, with
cross-channel coupling occurring implicitly through the fixed
orthogonal matrices $\Phi_C$ and $\Phi_C^\top$ at the encoder
boundaries. This algebraic structure is analogous to group
convolutions, where independent filter groups process channel subsets.
The key difference is that the c-product coupling is \emph{exact} and
\emph{parameter-free}: the $\Phi_C / \Phi_C^\top$ pair ensures that
inter-channel information is preserved globally despite per-slice
independence, without introducing any additional learnable parameters.

	\section{Numerical Experiments}
	\label{sec:experiments}
	
	The primary contribution of this work is the algebraic framework 
	developed in Sections~\ref{sec:cosine_product}--\ref{sec:analysis}. 
	The experiments in this section serve as a \emph{proof of concept}, 
	with two objectives,  verify that the theoretical $1/C$ 
	parameter reduction (Eq.~\eqref{eq:param_ratio}) is realized in 
	practice, and  assess whether the tensorized model retains 
	competitive accuracy despite the structural constraints imposed by 
	the block-diagonal c-product formulation. We evaluate on 
	classification and segmentation tasks using controlled, small-scale 
	configurations that isolate the effect of the c-product 
	tensorization. Large-scale validation (e.g., ImageNet, deeper 
	architectures) is left to future work.

	An important methodological point must be emphasized. The c-product
	tensorization is not a specific architecture: it is a \emph{general algebraic
		strategy} that can be applied to any Transformer-based model. Given any
	standard Transformer operating on flattened tokens
	$X \in \mathbb{R}^{N \times dC}$, the c-product reformulation replaces every
	linear projection  by the tensor cosine product
	  on the unflattened representation
	$\mathcal{X} \in \mathbb{R}^{N \times d \times C}$, yielding a $1/C$
	parameter reduction with no other architectural change. This strategy applies
	identically to DeiT, Swin, or any other Transformer variant.
	
	For this reason, comparing TCP-ViT against a \emph{different} architecture
	(e.g., DeiT or Swin) would conflate two independent factors: the base
	architecture and the tensorization strategy. The scientifically rigorous
	comparison is between a given Transformer and its c-product counterpart,
	with all other design choices held constant. This is exactly the protocol we
	adopt.
	
	Concretely, we define \textbf{Std-ViT} as a vanilla Vision Transformer with
	the same depth $L$, number of heads $H$, expansion ratio $r_{\mathrm{ff}}$,
	and training protocol as TCP-ViT. The \emph{only} difference is the token
representation: Std-ViT flattens each patch to
$\mathbf{x}_i \in \mathbb{R}^{d_{\mathrm{eff}}}$ with
$d_{\mathrm{eff}} = P^2 C$, while TCP-ViT represents each patch as a tensor
$\mathcal{X}_i \in \mathbb{R}^{d \times C}$ and processes it via the
c-product $\star_c$.
Any observed performance difference therefore isolates the
effect of the c-product tensorization.

	\subsection{Image Classification}
	\label{sec:classification}
	
	{Datasets.}
	We evaluate on three standard benchmarks:
	\begin{itemize}[leftmargin=2em]
		\item \textbf{CIFAR-10}~\cite{krizhevsky2009learning}: 60{,}000 colour
		images ($32 \times 32$) across 10 categories (50K train / 10K test).
		\item \textbf{SVHN}~\cite{netzer2011}: over 600{,}000 digit images
		($32 \times 32$) from Google Street View (10 classes).
		\item \textbf{STL-10}~\cite{coates2011analysis}: 13{,}000 labelled
		images ($96 \times 96$, resized to $32 \times 32$) across 10 classes
		(5K train / 8K test).
	\end{itemize}
	To validate architectural properties under controlled conditions, both models
	are also trained on identically subsampled subsets: 10{,}000 training and
	2{,}000 test samples per dataset.
	
	\nd {\bf Architecture.}
	Both models process $32 \times 32 \times 3$ images with patch size $P = 4$,
	yielding $N = 64$ patches. Each patch is a tensor
	$\mathcal{X}_i \in \mathbb{R}^{16 \times 3}$.
	\begin{itemize}[leftmargin=2em]
		\item \textbf{Std-ViT} flattens each patch to $d_{\mathrm{eff}} = P^2 C = 48$ and processes through a single encoder.
		\item \textbf{TCP-ViT} represents each patch as
$\mathcal{X}_i \in \mathbb{R}^{d \times C}$ with $d = P^2 = 16$ and
$C = 3$, and processes it via the c-product $\star_c$.
	\end{itemize}
	Both use $L = 4$ layers, $H = 4$ heads, and expansion ratio
	$r_{\mathrm{ff}} = 4$. This yields 119{,}194 parameters for Std-ViT and
	43{,}114 for TCP-ViT, a ratio of $0.362\times$ (theoretical: $1/C = 0.333\times$).
	\\
	Both models are trained for 150 epochs (subsampled) or 200 epochs (full) using
	AdamW~\cite{loshchilov2019decoupled} (lr $= 0.01$, weight decay $= 0.01$) with
	cosine annealing, batch size~256, gradient clipping at norm~1.0, and
	mixed-precision training. Data augmentation consists of random cropping
	(padding~4) and random horizontal flipping. Images are normalized using
	per-dataset statistics for CIFAR-10 and ImageNet statistics for SVHN and STL-10.
	
	We report top-1 classification accuracy on the test set. Throughout all
	experimental tables, $\Delta$ denotes the absolute performance difference
	\begin{equation}\label{eq:delta_def}
		\Delta = \mathrm{metric}(\text{TCP-ViT}) - \mathrm{metric}(\text{Std-ViT}),
	\end{equation}
	expressed in percentage points. A positive $\Delta$ indicates that TCP-ViT
	outperforms Std-ViT; a negative $\Delta$ indicates a performance loss.
	Best results per row are shown in \textbf{bold}.

	
	\begin{table}[htbp]
		\centering
		\caption{Classification accuracy on subsampled datasets (10K train / 2K test).
			TCP-ViT uses $0.362\times$ the parameters of Std-ViT (theoretical: $1/C = 0.333\times$).}
		\label{tab:cls_sub}
		\begin{tabular}{lccccr}
			\toprule
			\textbf{Dataset} & \textbf{Std-ViT (\%)} & \textbf{TCP-ViT (\%)} & $\boldsymbol{\Delta}$ & \textbf{Std Params} & \textbf{TCP-ViT Params} \\
			\midrule
			CIFAR-10 & 61.8 & \textbf{63.6} & $+1.8$  & 119{,}194 & 43{,}114 \\
			SVHN     & 49.8 & \textbf{63.8} & $+14.0$ & 119{,}194 & 43{,}114 \\
			STL-10   & 52.5 & \textbf{55.9} & $+3.4$  & 119{,}194 & 43{,}114 \\
			\midrule
			\multicolumn{4}{l}{Parameter ratio} & \multicolumn{2}{r}{$0.362\times$ (theoretical: $0.333\times$)} \\
			\bottomrule
		\end{tabular}
	\end{table}
	
	Table~\ref{tab:cls_sub} reports the results. Under the subsampled protocol,
	TCP-ViT matches or exceeds Std-ViT on all three datasets: $+1.8\%$ on CIFAR-10,
	$+14.0\%$ on SVHN, and $+3.4\%$ on STL-10. These gains are attributed to
	implicit regularization: Std-ViT, with $2.76\times$ more parameters, exhibits
	significant overfitting (training loss near zero, diverging validation loss),
	while TCP-ViT's reduced capacity constrains the model and yields a more favorable
	bias--variance trade-off under limited data.
	
	The large gain on SVHN ($+14.0\%$) warrants particular attention. SVHN images
	contain cluttered backgrounds and variable lighting, which amplify overfitting
	in the over-parameterized Std-ViT. 

	
	\begin{table}[htbp]
		\centering
		\caption{Classification accuracy on full CIFAR-10 (50K train / 10K test,
			200 epochs). TCP-ViT retains 93.3\% of performance with 36.2\% of the parameters.}
		\label{tab:cls_full}
		\begin{tabular}{lcccr}
			\toprule
			\textbf{Dataset} & \textbf{Std-ViT (\%)} & \textbf{TCP-ViT (\%)} & $\boldsymbol{\Delta}$ & \textbf{Param ratio} \\
			\midrule
			CIFAR-10 (full) & 78.7 & 73.4 & $-5.3$ & $0.362\times$ \\
			\bottomrule
		\end{tabular}
	\end{table}
	
	Table~\ref{tab:cls_full} reports the full-scale result. On the complete
	CIFAR-10, Std-ViT achieves 78.7\% while TCP-ViT reaches 73.4\%
	($\Delta = -5.3\%$). This represents \textbf{93.3\% performance retention with
		only 36.2\% of the parameters}, a trade-off that is consistent with standard
	model compression methods, which typically report 3--8\% accuracy degradation
	for $2$--$3\times$ compression. The accuracy gap confirms that when sufficient
	training data is available, the strict block-diagonal structure in the DCT
	domain may discard useful cross-frequency interactions that a dense layer can
	capture.
	\\
	The c-product decomposes the channel dimension using the DCT, replacing the
	$P^2 C$-dimensional joint representation with $C$ independent
	$P^2$-dimensional representations. Each frontal slice captures a distinct
	frequency component: slice $k{=}1$ corresponds to the channel mean
	(low-frequency), while subsequent slices encode inter-channel contrasts
	(higher-frequency). This preserves discriminative features while eliminating
	the redundancy of the standard flattening $\mathrm{flatten}(P^2, C) \to \mathbb{R}^{P^2 C}$.

	\subsection{Semantic Segmentation}
	\label{sec:segmentation}
	
	We evaluate on the Oxford-IIIT Pet dataset~\cite{parkhi2012cats}, which
	contains 7{,}349 images of 37 cat and dog breeds with pixel-level trimap
	annotations. Following standard practice, we convert to binary segmentation
	(foreground vs.\ background), discarding boundary pixels (label~255). The
	dataset is split 80/20 into training (5{,}879) and validation (1{,}470) sets.
	\\
	Both models share an identical CNN decoder (4 upsampling stages). The backbone
	processes $128 \times 128$ images with patch size $P = 8$, yielding $N = 256$
	patches. Std-ViT uses $d_{\mathrm{eff}} = P^2 C = 192$; TCP-ViT operates on
	$C = 3$ slices of dimension $d = 64$. Both use $L = 4$ layers, $H = 4$ heads,
	$r_{\mathrm{ff}} = 2$. After IDCT reconstruction, TCP-ViT outputs features of
	dimension $dC = 192$, matching the Std-ViT decoder input.
	
	\nd {\bf Training protocol.} We used: 150 epochs, AdamW (lr $= 5 \times 10^{-4}$, weight decay $0.01$), cosine
	annealing, cross-entropy loss (ignore index~255), batch size~16, gradient
	clipping at norm~1.0, mixed-precision. Input resized to $128 \times 128$ with
	ImageNet normalization. Best validation mIoU reported.
	\\
	Mean IoU (mIoU), mean Dice (mDice), pixel accuracy (PA), and per-class IoU/Dice.

	
	\begin{table}[htbp]
		\centering
		\caption{Parameter breakdown for segmentation. The Transformer encoder achieves
			$0.338\times$, closely matching the theoretical $1/C = 0.333\times$.}
		\label{tab:seg_params}
		\begin{tabular}{lrrr}
			\toprule
			Component & Std-ViT & TCP-ViT & Ratio \\
			\midrule
			\textbf{Total} & 1{,}570{,}978 & 751{,}586 & $0.478\times$ \\
			\quad Backbone & 1{,}275{,}072 & 455{,}680 & $0.357\times$ \\
			\quad\quad Transformer Encoder(s) & 1{,}188{,}480 & 402{,}048 & $0.338\times$ \\
			\quad\quad Embeddings \& Tokens & 86{,}592 & 53{,}632 & $0.619\times$ \\
			\quad CNN Decoder & 295{,}906 & 295{,}906 & $1.000\times$ \\
			\bottomrule
		\end{tabular}
	\end{table}
	
\begin{figure}[htbp]
		\centering
		\includegraphics[width=\linewidth]{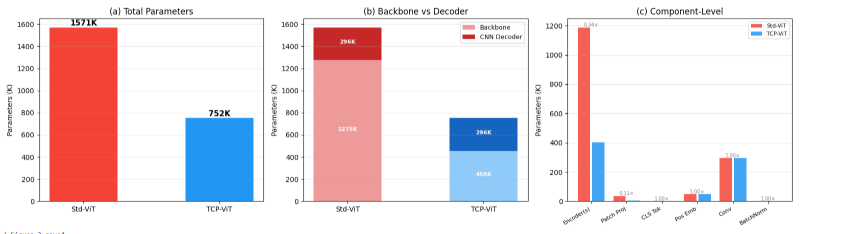}
		\caption{Parameter distribution. (a)~Total parameters: TCP-ViT uses $0.478\times$ of Std-ViT. (b)~Stacked breakdown. (c)~Component-level comparison.}
		\label{fig:seg_params}
	\end{figure}

	Table~\ref{tab:seg_params} presents the parameter breakdown. The Transformer
	encoder achieves a compression ratio of $0.338\times$, closely matching the
	theoretical $1/C = 1/3 \approx 0.333\times$ from
	Eq.~\eqref{eq:param_ratio}. The slight deviation arises from bias terms and
	LayerNorm parameters, which scale linearly. The CNN decoder is identical
	($1.000\times$), confirming it operates on reconstructed features after IDCT.
	Figure~\ref{fig:seg_params} provides a visual comparison.
	
	\begin{table}[htbp]
		\centering
		\caption{Segmentation performance on Oxford-IIIT Pet. TCP-ViT retains 98.9\%
			of the mIoU with $2.09\times$ fewer parameters. $\Delta$ denotes the absolute difference (TCP-ViT $-$ Std-ViT).}
		\label{tab:seg_performance}
		\begin{tabular}{lrrr}
			\toprule
			Metric & Std-ViT & TCP-ViT & $\Delta$ \\
			\midrule
			mIoU (\%)              & 87.8 & 86.8 & $-1.0$ \\
			mDice (\%)             & 93.5 & 92.9 & $-0.6$ \\
			Pixel Acc.\ (\%)       & 94.2 & 93.6 & $-0.6$ \\
			\midrule
			IoU: Background        & 84.0 & 82.7 & $-1.3$ \\
			IoU: Foreground (Pet)  & 91.7 & 90.9 & $-0.8$ \\
			\midrule
			Dice: Background       & 91.3 & 90.5 & $-0.8$ \\
			Dice: Foreground (Pet) & 95.7 & 95.2 & $-0.4$ \\
			\midrule
			Parameters & 1{,}570{,}978 & 751{,}586 & $0.478\times$ \\
			\bottomrule
		\end{tabular}
	\end{table}
	
\begin{figure}[htbp]
		\centering
		\includegraphics[width=\linewidth]{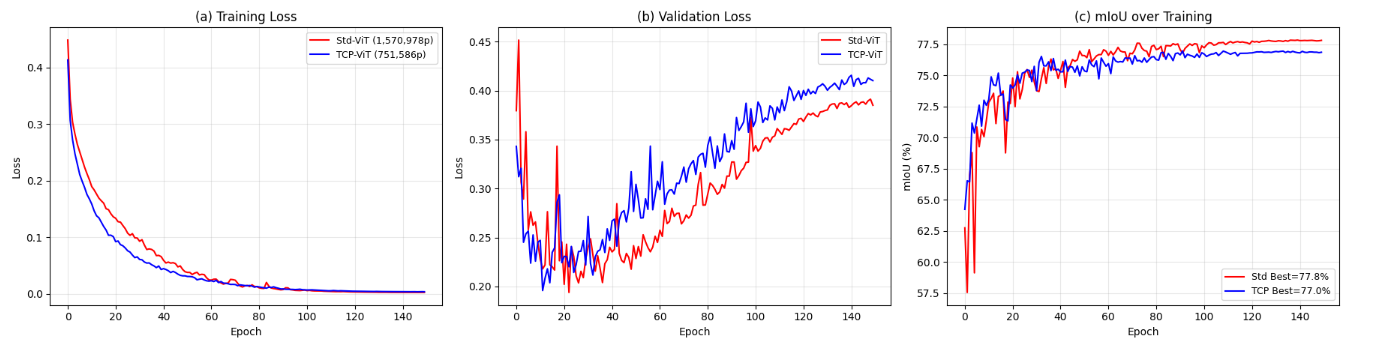}
		\caption{Training dynamics. (a)~Training loss, (b)~validation loss, (c)~mIoU. TCP-ViT (blue) closely tracks Std-ViT (red) with a stable ${\sim}1\%$ gap.}
		\label{fig:seg_curves}
	\end{figure}

	Table~\ref{tab:seg_performance} summarizes the results. TCP-ViT achieves 86.8\%
	mIoU versus 87.8\% for Std-ViT ($\Delta = -1.0\%$) with $2.09\times$ fewer
	parameters. The degradation is small and consistent across all metrics. The gap
	is smaller on the foreground class ($\Delta_{\mathrm{IoU}} = -0.8\%$) than on
	the background ($\Delta_{\mathrm{IoU}} = -1.3\%$), suggesting that the
	c-product processing preserves object-level semantics effectively while losing
	marginal precision on less structured regions.
	\\	
	Figure~\ref{fig:seg_curves} shows the training dynamics over 150 epochs. Both
	models converge at comparable rates, with TCP-ViT exhibiting slightly higher
	losses. The mIoU gap stabilizes around 1\% after epoch~50. Neither model shows
	significant overfitting.
	
\begin{figure}[htbp]
		\centering
		\includegraphics[width=0.85\linewidth]{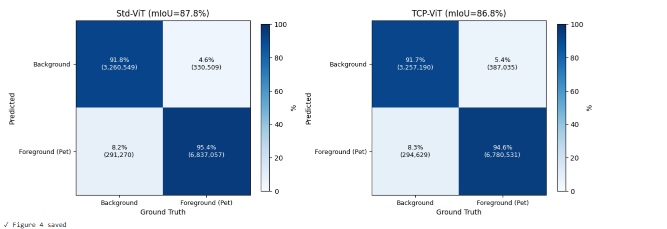}
		\caption{Normalized confusion matrices. Both models show strong diagonal dominance.}
		\label{fig:seg_confusion}
	\end{figure}

	\nd Figure~\ref{fig:seg_confusion} shows the normalized confusion matrices. Both
	models achieve high true positive rates ($>$90\% foreground, $>$82\%
	background).
	
\begin{figure*}[htbp]
		\centering
		\includegraphics[width=\textwidth]{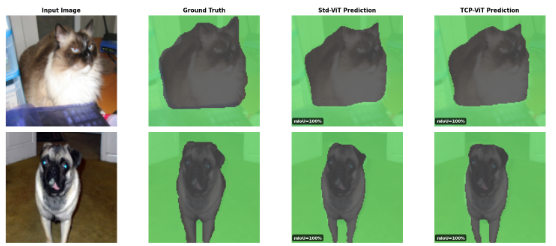}
		\caption{Qualitative results. Columns: input, ground truth, Std-ViT, TCP-ViT. Green overlay: foreground. Per-image mIoU shown on each prediction.}
		\label{fig:seg_visual}
	\end{figure*}

	\nd Figure~\ref{fig:seg_visual} presents predictions on validation images. Both
	models produce visually similar masks. TCP-ViT occasionally shows slightly less
	precise boundaries in regions with complex backgrounds.
	
\begin{figure*}[htbp]
		\centering
		\includegraphics[width=\textwidth]{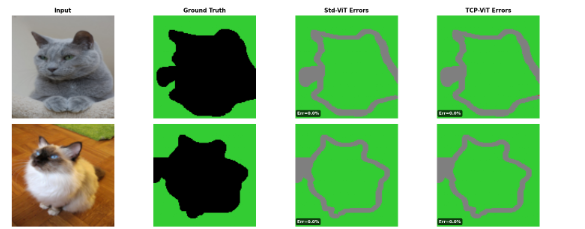}
		\caption{Error maps. Green = correct, red = error, gray = ignored. Errors are concentrated at boundaries for both models.}
		\label{fig:seg_errors}
	\end{figure*}

	\nd Figure~\ref{fig:seg_errors} visualizes error maps. Errors are concentrated
	along object boundaries for both models, with TCP-ViT showing marginally more
	boundary errors.
	
\begin{figure}[htbp]
		\centering
		\includegraphics[width=0.75\linewidth]{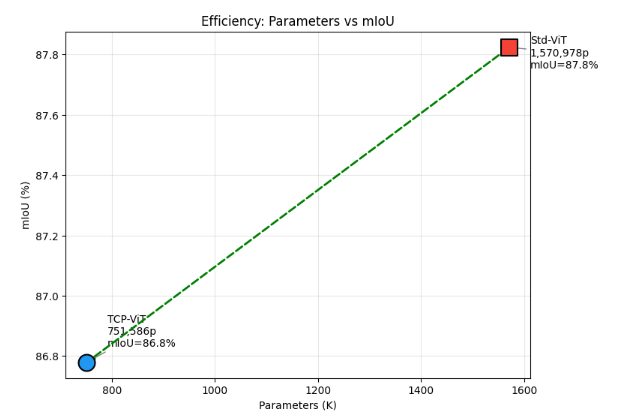}
		\caption{Efficiency trade-off: parameters vs.\ mIoU. The dashed arrow indicates the compression direction.}
		\label{fig:seg_efficiency}
	\end{figure}

	\nd Figure~\ref{fig:seg_efficiency} presents the parameter--performance trade-off.
	TCP-ViT reduces parameters by $2.09\times$ while sacrificing only 1.0\% mIoU,
	yielding a compression efficiency of
	$\eta = \Delta\mathrm{mIoU} / \Delta\mathrm{params} = 1.0\% / 52.2\% \approx 0.019$,
	i.e., less than 0.02\% mIoU lost per 1\% parameter reduction.
	
	\subsection{Discussion}
	\label{sec:discussion}
	
	We synthesize the experimental findings along four axes: parameter efficiency,
	data regime sensitivity, task generality, and architectural interpretation.
	\\
	The theoretical analysis of Section~\ref{sec:param_efficiency} predicts a
	uniform $1/C$ parameter reduction in all linear projections. The experiments
	confirm this prediction closely: the Transformer encoder compression ratio is
	$0.362\times$ for classification and $0.338\times$ for segmentation, both
	near the theoretical $1/C = 1/3 \approx 0.333\times$
	(Table~\ref{tab:param_general} and Table~\ref{tab:seg_params}). The small
	deviations are fully accounted for by bias terms and LayerNorm parameters,
	which scale linearly in $dC$ and thus do not benefit from the $1/C$ reduction.
	When task-specific components are shared (e.g., the CNN decoder in
	segmentation), the \emph{overall} compression ratio is diluted to
	$0.478\times$, but the Transformer backbone ratio remains near its
	theoretical value. This confirms that the c-product reduction is exact and
	architecture-agnostic within the encoder.
	\\
	The results reveal a clear interaction between model capacity and data
	availability. Under the subsampled protocol (10K training samples),
	TCP-ViT \emph{outperforms} Std-ViT on all three datasets
	($\Delta = +1.8\%$ on CIFAR-10, $+14.0\%$ on SVHN, $+3.4\%$ on STL-10;
	Table~\ref{tab:cls_sub}). This is consistent with classical bias--variance
	analysis: the over-parameterized Std-ViT (119K parameters for 10K samples)
	overfits severely, while TCP-ViT's reduced capacity acts as an implicit
	regularizer, constraining the hypothesis space and improving generalization.
	The effect is most pronounced on SVHN, where cluttered backgrounds and
	variable lighting amplify overfitting. By contrast, on the full CIFAR-10
	(50K samples), Std-ViT achieves 78.7\% versus 73.4\% for TCP-ViT
	($\Delta = -5.3\%$; Table~\ref{tab:cls_full}). With sufficient data, the
	additional $2.76\times$ parameters of Std-ViT enable it to capture
	cross-frequency interactions that the block-diagonal c-product structure
	cannot represent. Nevertheless, TCP-ViT retains 93.3\% of the accuracy with
	only 36.2\% of the parameters, a trade-off comparable to standard compression
	methods that typically report 3--8\% degradation for $2$--$3\times$
	compression.
	
	\nd {\bf Task generality: classification vs.\ dense prediction.}
	The segmentation experiment demonstrates that the c-product framework
	extends beyond classification to dense prediction. On the Oxford-IIIT Pet
	benchmark, TCP-ViT achieves 86.8\% mIoU versus 87.8\% for Std-ViT
	($\Delta = -1.0\%$; Table~\ref{tab:seg_performance}), retaining 98.9\% of
	the performance. The accuracy gap is notably smaller than for full-scale
	classification ($-1.0\%$ vs.\ $-5.3\%$), which can be attributed to two
	factors: (i)~the segmentation task operates at higher resolution
	($128 \times 128$ vs.\ $32 \times 32$), increasing the token count and
	thus the relative benefit of parameter efficiency; and (ii)~the binary
	segmentation objective may be less sensitive to fine-grained cross-frequency
	interactions than 10-class classification. The error analysis
	(Figure~\ref{fig:seg_errors}) reveals that degradation is concentrated at
	object boundaries, consistent with the loss of inter-slice coupling in the
	DCT domain.
	
	\nd The c-product imposes a \emph{block-diagonal} structure in the DCT domain:
	each of the $C$ frontal slices is processed by an independent linear map,
	with cross-channel coupling handled exclusively by the fixed orthogonal
	matrices $\Phi_C$ and $\Phi_C^\top$. This structure has both advantages and
	limitations. On the positive side, it provides a principled and exact
	(lossless) parameter reduction, preserved gradient norms due to
	orthogonality, and a natural decomposition into interpretable frequency
	components (slice $k{=}1$ captures the channel mean, higher slices capture
	inter-channel contrasts). On the negative side, the strict block-diagonal
	constraint prevents the model from learning arbitrary cross-frequency
	interactions, which limits expressiveness when sufficient data is available.
	This trade-off is analogous to group convolutions, where splitting channels
	into independent groups reduces parameters at the cost of inter-group
	communication.
	
	\nd {\bf Limitations and future directions.}
	Several limitations of the current work should be acknowledged.
	\begin{enumerate}[leftmargin=2em]
		\item \emph{Sequential implementation.} The current  implementation
		processes the $C$ frontal slices sequentially. A parallel implementation   or multi-stream CUDA kernels
		is required to realize wall-clock speedups proportional to the theoretical
		$1/C$ FLOPs ratio.
		
		\item \emph{Task head dilution.} Shared task-specific components (e.g.,
		the CNN decoder in segmentation) are not tensorized and therefore dilute
		the overall compression from $1/C$ to a higher value. Tensorizing the
		decoder is a natural extension.
		
		\item \emph{Accuracy gap on large datasets.} The $-5.3\%$ gap on
		full-scale CIFAR-10 suggests that the strict block-diagonal structure
		discards useful cross-frequency interactions. Future work could explore
		learnable inter-slice coupling (e.g., a lightweight mixing layer between
		slices) or partial c-product formulations that relax the block-diagonal
		constraint while preserving most of the parameter savings.
		
		\item \emph{Scale of experiments.} The current evaluation uses small-scale
		ViT configurations ($L{=}4$, $d_{\mathrm{eff}} \leq 192$) on
		low-resolution benchmarks ($32 \times 32$ to $128 \times 128$).
		Validation on larger-scale settings (e.g., ImageNet at $224 \times 224$
		with deeper architectures) is needed to confirm the scalability of the
		c-product framework.
	\end{enumerate}

	\section{Conclusion}
	\label{sec:conclusion}
	
	We developed a rigorous algebraic framework for tensorizing Vision 
	Transformers based on the tensor cosine product. The central 
	theoretical result is that replacing every linear projection in a 
	standard Transformer by the c-product   yields a provable and uniform $1/C$ parameter 
	reduction that is exact (lossless), architecture-agnostic, and 
	independent of the network depth, number of heads, or FFN expansion 
	ratio. The reduction arises from the block-diagonal structure 
	induced by the fixed orthogonal DCT matrix $\Phi_C$, which handles 
	cross-channel coupling at zero parametric cost while preserving 
	gradient norms. 
	Proof-of-concept experiments on classification (CIFAR-10, SVHN, 
	STL-10) and segmentation (Oxford-IIIT Pet) benchmarks confirmed 
	that the theoretical compression ratios are achieved in 
	practice ($0.362\times$ and $0.338\times$ vs.\ the theoretical 
	$0.333\times$), the tensorized model retains competitive 
	accuracy, and the c-product acts as an implicit regularizer 
	in low-data regimes. These results validate the algebraic 
	framework and motivate further investigation.
	\\
	Several directions remain open. On the theoretical side, 
extensions to learnable inter-slice coupling and partial c-product 
formulations could relax the block-diagonal constraint while 
preserving most of the parameter savings. On the practical side, 
parallel GPU implementations, large-scale validation on ImageNet 
with deeper architectures, application to other Transformer 
variants (Swin, DeiT), and extensions to higher-order 
tensors for video and volumetric data constitute natural next 
steps. The framework is particularly promising for hyperspectral 
imaging, where $C$ ranges from tens to hundreds of spectral bands 
and the $1/C$ reduction becomes correspondingly more significant.


\end{document}